%% file: main.tex
\documentclass[conference,a4paper]{IEEEtran}
\IEEEoverridecommandlockouts
\usepackage{cite}
\usepackage{amsmath,amssymb,amsfonts}
\usepackage{algorithmic}
\usepackage{graphicx}
\usepackage{textcomp}
\usepackage{xcolor}
\usepackage[utf8]{inputenc} % allow utf-8 input
\usepackage[T1]{fontenc}    % use 8-bit T1 fonts
\usepackage{hyperref}       % hyperlinks
\usepackage{url}            % simple URL typesetting
\usepackage{booktabs}       % professional-quality tables
\usepackage{amsfonts}       % blackboard math symbols
\usepackage{nicefrac}       % compact symbols for 1/2, etc.

\usepackage{graphicx} % includegraphics, figures, etc
\usepackage{caption}
\usepackage{subcaption} % for subfigures
\usepackage{microtype}      % microtypography
\usepackage{multirow, makecell} % makecell allows line break in cell
\usepackage{hyperref}       % hyperlinks
\usepackage[normalem]{ulem}
\usepackage{comment}

\def\BibTeX{{\rm B\kern-.05em{\sc i\kern-.025em b}\kern-.08em
    T\kern-.1667em\lower.7ex\hbox{E}\kern-.125emX}}

\begin{document}

\title{Classifying Breast Histopathology Images with a\\ Ductal Instance-Oriented Pipeline
\thanks{
Research reported in this article was supported by grants R01 CA172343, R01 CA140560, U01CA231782, and R01 CA200690 from the National Cancer Institute of the National Institutes of Health.
}
}

\author{\IEEEauthorblockN{
Beibin Li\IEEEauthorrefmark{1}\IEEEauthorrefmark{3},
Ezgi Mercan\IEEEauthorrefmark{3},
Sachin Mehta\IEEEauthorrefmark{1}, \\
Stevan Knezevich\IEEEauthorrefmark{6},
Corey W. Arnold\IEEEauthorrefmark{2},
Donald L. Weaver\IEEEauthorrefmark{4},
Joann G. Elmore\IEEEauthorrefmark{2},
Linda G. Shapiro\IEEEauthorrefmark{1}\\ \\
\IEEEauthorrefmark{1}University of Washington, Seattle, WA
\IEEEauthorrefmark{2}University of California, Los Angles, CA\\
\IEEEauthorrefmark{3}Seattle Children's Hospital, Seattle, WA
\IEEEauthorrefmark{4}University of Vermont, Burlington, VT\\
\IEEEauthorrefmark{6}Pathology Associates, Clovis, CA
}}

\maketitle

\begin{abstract}
In this study, we propose the Ductal Instance-Oriented Pipeline (DIOP) that contains a duct-level instance segmentation model, a tissue-level semantic segmentation model, and three-levels of features for diagnostic classification. Based on recent advancements in instance segmentation and the Mask R-CNN model, our duct-level segmenter tries to identify each ductal individual inside a microscopic image; then, it extracts tissue-level information from the identified ductal instances. Leveraging three levels of information obtained from these ductal instances and also the histopathology image, the proposed DIOP outperforms previous approaches (both feature-based and CNN-based) in all diagnostic tasks; for the four-way classification task, the DIOP achieves comparable performance to general pathologists in this unique dataset. The proposed DIOP only takes a few seconds to run in the inference time, which could be used interactively on most modern computers. More clinical explorations are needed to study the robustness and generalizability of this system in the future.
\end{abstract}


\section{Introduction}

Breast cancer is one of the most common cancers for females: about 13\% of women will develop breast cancer over their lifetimes, and 2.6\% of women will die from breast cancer in the United States \cite{breast2019facts}.
The recent development of Artificial Intelligence (AI) screening tools for mammography \cite{mckinney2020international} could reduce the second reader's workload, but diagnosing breast cancer is still a time-consuming and challenging task.
Physicians usually recommend  breast biopsies for diagnosis and for the development of treatment plans after finding suspicious areas in a mammogram, ultrasound, or magnetic resonance imaging (MRI).
% Analyzing breast biopsies is the most important step for breast cancer diagnosis, but even pathologists only agree with each other in 70\% of cases. 

When analyzing breast biopsies, pathologists usually focus on ducts, because most breast cancers begin in the terminal ducts or lobules of the breast \cite{breast2019facts}.
In ductal carcinoma in situ (DCIS), cells inside ducts undergo malignant transformation to cancer cells;
in invasive breast cancer, these abnormal cells have escaped from the duct and are growing in the surrounding tissue \cite{breast2019facts}.
Hence, tissues surrounding duct borders are the most relevant regions for pathologists and also machine learning models to focus on. 

\begin{figure}[t]
    \centering
    \begin{subfigure}[b]{0.11\textwidth}
        \includegraphics[width=0.99\textwidth]{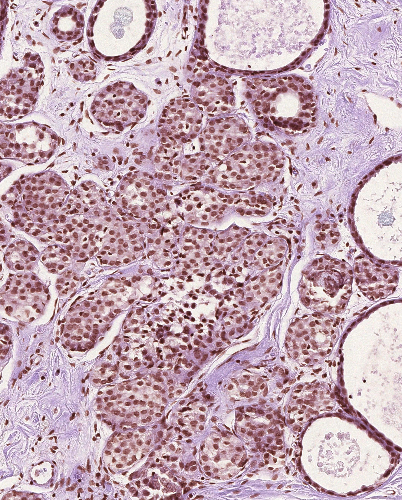}
        \caption{ROI image}
        \label{fig:instance:input}
    \end{subfigure}
    \begin{subfigure}[b]{0.11\textwidth}
        \includegraphics[width=0.99\textwidth]{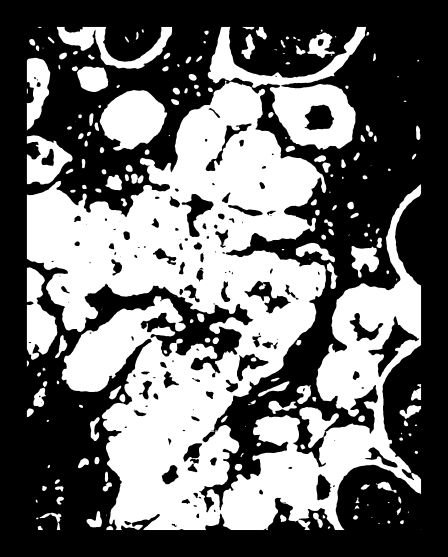}
        \caption{Binary}
        \label{fig:instance:binary}
    \end{subfigure}
    \begin{subfigure}[b]{0.11\textwidth}
        \includegraphics[width=0.99\textwidth]{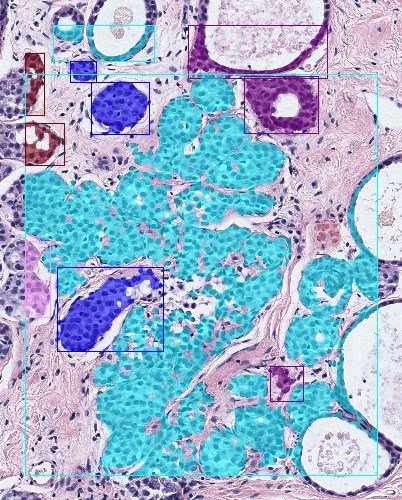}
        \caption{Previous}
        \label{fig:instance:connected}
    \end{subfigure}
    \begin{subfigure}[b]{0.11\textwidth}
        \includegraphics[width=0.99\textwidth]{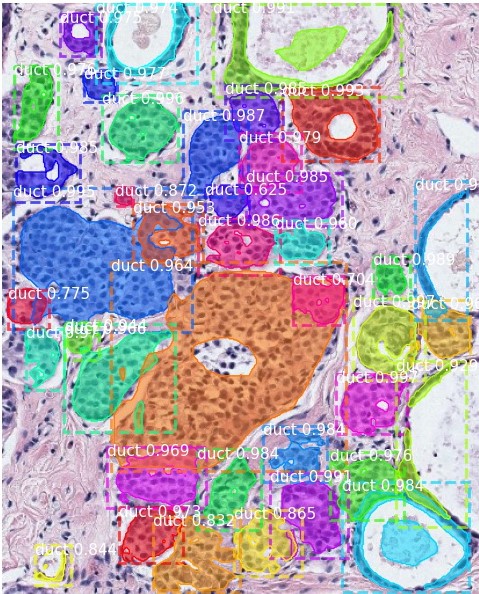}
        \caption{Ours}
        \label{fig:instance:ours}
    \end{subfigure}
  \caption{\textbf{Duct instances:} (a) the input ROI image in RGB color space; (b) the binary image inferred from tissue-level semantic segmentation, where the white pixels are ducts; (c) duct instances found by mathematical morphology and connected component algorithm; (d) the ducts inferred from our system. In (c) and (d), each color represents one duct instance.
  The connected component method (c) could not distinguish duct instance from the conglomerated region, even if it has been used to solve similar problems (e.g. in \cite{mercan2019assessment,li2018path}, etc.).
  }
  \label{fig:instance_compare}
\end{figure}

Based on the importance of ductal regions, Mercan et al. designed structure features \cite{mercan2019assessment} to summarize the architectural characteristics in duct-based structures.
Their method can emulate pathologists' behaviors to interpret diagnostic decisions and has been shown to outperform pathologists on the difficult task of categorically differentiating DCIS from Atypia.
They identified duct instances (i.e. all pixels that are part of an individual breast duct or lobule) by applying a union-find algorithm to split semantic segmentation predictions of specific tissue classes into smaller duct regions. 
However, many ducts are entangled in breast biopsies, and this approach cannot distinguish a duct instance that is adjacent to other ducts, as shown in Figure \ref{fig:instance_compare}.
Moreover, extracting their structure features from biopsies can take hours on a computer, because this algorithm is not suitable for parallel processing on multiple computer cores.
The accuracy and computational requirements need improvement if the goal is to create an interactive and real-time tool for clinical use.

On the other hand, many research groups have designed end-to-end deep learning systems to classify breast histopathology images, including fully-convolutional networks (FCNs) \cite{gecer2018detection}, patch- to ROI-level feature representation \cite{mercan2019patch}, and graph convolutional networks \cite{aygunecs2020graph}. 
Mehta et al. also designed Y-Net \cite{mehta2018net}, which can perform image segmentation and diagnostic classification at the same time. 
These systems are usually accurate and fast because of the recent advancement of Graphical Processing Units (GPUs) and parallelism for matrix manipulation.
Nevertheless, these methods are somewhat blind to the underlying pathological and structural features that led to the clinical diagnosis. They only provide heat maps, attention maps, or patches for visualization; they do not guarantee a focus on ductal regions and cannot offer a decisive interpretation as \cite{mercan2019assessment} does for pathologists. 
Knowing why an AI algorithm makes a certain diagnostic decision is crucial in clinical practice.

Researchers often have to balance the trade-offs between speed, accuracy, and interpretability within these computer-aided diagnosis (CAD) tools.
In this study, we created a Ductal Instance-Oriented Pipeline (DIOP) that can identify individual duct structures in breast biopsies, extract features, and classify breast cancer diagnoses. The proposed pipeline improves upon previous approaches on all diagnosis tasks, outperforms human general pathologists in two out of three binary classification tasks, and achieves comparable performance to general pathologists in a four-way diagnostic classification task, distinguishing among Benign, Atypia, Ductal Carcinoma in Situ, and Invasive Cancer category examples.

\section{Background}

Recent developments in breast cancer assessment, semantic segmentation, instance segmentation, and weakly supervised learning for medical imaging provide the groundwork for our study.
Semantic segmentation is a common task in medical imaging; it partitions an image into multiple tissues, grades, or classes by classifying each pixel inside the image.
For example, LSBB \cite{mehta2018learning}, Y-Net \cite{mehta2018net}, and ESPNet \cite{mehta2018espnet} were designed for breast biopsy semantic segmentation;
multi-scale U-Net \cite{li2017multi}, vanilla FCNs \cite{ing2018semantic}, EM-based models \cite{li2018based}, and attention models \cite{li2019attention} were  created for prostate cancer;  specialized auto-encoder \cite{attia2017skin}, FCNs \cite{goyal2017multi}, and U-Net \cite{van2020segmentation} were used for  melanoma segmentation tasks.
Unfortunately, these methods could not identify duct instances inside a region of interest (ROI), because semantic segmentation could not differentiate these instances from the semantic labels. 
Instead, instance segmentation labels, which contain all pixels that are part of each breast duct or lobule, are needed.

While semantic segmentation has been widely applied to cancer diagnosis, instance segmentation is rarely used.
Li et al. designed Path R-CNN (regions with convolutional neural network features) \cite{li2018path} based on Mask R-CNN \cite{he2017mask} to classify glands and grade prostate cancer, which pioneered instance segmentation for medical imaging. 
In the recent two years, instance segmentation for nuclei \cite{zhang2018nuclei}, cluttered cells \cite{guerrero2018multiclass}, polyps \cite{kang2019ensemble}, and other tissues have been developed.
In breast biopsies, each individual duct might contain different structural information about tissues, which could not be provided by tissue-level semantic segmentation.
Hence, performing instance segmentation for ducts could be valuable for classifying histopathology images.

Acquiring instance segmentation labels for ducts is a tedious and time-consuming process, and no public datasets are available for instance segmentation on ducts to the best of our knowledge.
For images with simple structures, semantic segmentation labels can be easily converted to instance segmentation labels by using union-find, connected components, or other rule-based algorithms. 
Unfortunately, ducts inside of breast biopsies are too complex for these rule-based label conversion algorithms, and instance annotations are needed in order to train an instance segmenter.

Geometrically, the shape of a duct is similar to the shape of a "doughnut": it is usually ring-shaped with a circular thick border and central empty space, but often it does not have any holes because microscopic histopathology images are two-dimensional cross sections through three-dimensional structures. Breast ducts are analogous to pipes or tubes, and breast lobules are analogous to a hollow ball such as a tennis ball. Pre-cancerous conditions can also cause the lining of the ducts and lobules to proliferate and fill the holes. All these features complicate rule-based algorithms.  The difficulties compound when many ducts are adjacent to each other so that the borders of these ducts are difficult to distinguish. In DCIS and invasive cases, cancerous cells first begin to distort then escape from ducts, respectively, and thus duct cross sections develop into various and complex shapes.

Recent studies show that weak annotation \cite{zhou2018brief}, imperfect annotation \cite{tajbakhsh2020embracing}, active learning, and human-in-the-loop methods \cite{budd2019survey} can be used to alleviate this problem. These methods encouraged us to design an efficient annotation plan to find ducts inside breast biopsies.

\begin{figure*}[t!]
    \centering
    \includegraphics[width=\textwidth]{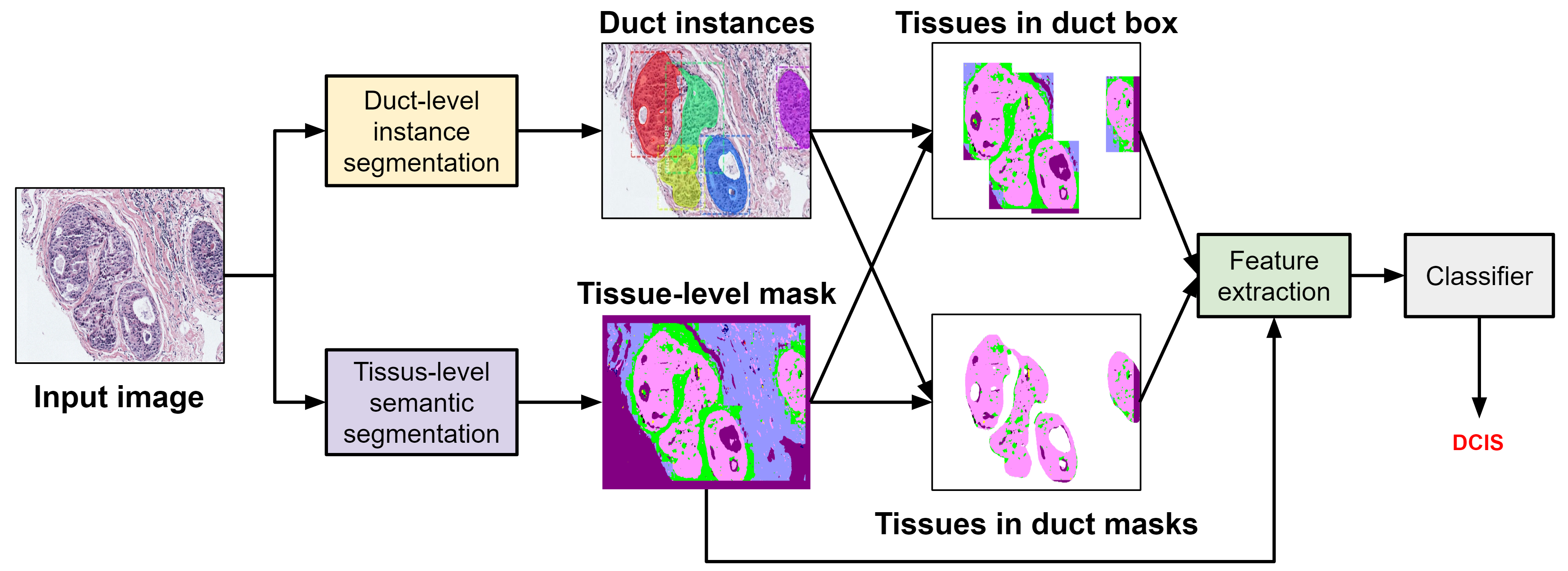}
   \caption{
   Ductal Instance-oriented Pipeline: this pipeline leverages existing tissue-level semantic segmentation, Mask R-CNN for duct-level instance segmentation, extracting features, and a classifier for diagnosis. 
   The feature extractors calculate the tissue histogram frequency and co-occurrence matrix for ROI-, box-, and mask-levels. 
   }
   \label{fig:duct_rcnn}
\end{figure*}

\section{System for Breast Cancer Diagnosis}
\label{sec:system}
Several clinical studies have shown that stromal tissues and ducts are important biomarkers for diagnosing breast cancer \cite{breast2019facts,bevers2009breast}. Motivated by these studies, we introduce a machine learning-based framework that accounts for these important bio-markers in cancer diagnosis. Our system consists of three components: (1) a duct-level instance segmentation model (Section \ref{sec:instance_segmentation}), (2) a tissue-level semantic segmentation model (Section \ref{sec:sem_seg}), and (3) a classifier with three-levels of extracted features (Section \ref{sec:pixel_features}). The input ROI is fed to both duct-level and tissue-level segmentation modules simultaneously to produce instances of ducts and tissue-level segmentation masks.  Histogram frequency and co-occurrence features are extracted at three different levels, from the duct, from the bounding boxes, and from the entire region of interest to predict the diagnosis. Our experimental results show that the proposed framework outperforms the previous state-of-the-art methods and matches the performance of pathologists.

\subsection{Duct-level instance segmentation}
\label{sec:instance_segmentation}
The instance segmentation network adopts the same structure as Mask R-CNN \cite{he2017mask}. The network consists of two stages. The first stage takes an ROI as an input and produces duct candidates. In the second stage, these candidates are then classified as duct or not. In addition to this classification, the second stage also produces bounding box coordinates as well as a pixel-wise mask of the duct. 

As in any other medical imaging task, collecting data is difficult because it requires experts to annotate data. To generate duct-level instance annotations, a weakly-supervised annotation framework is introduced that combines an annotator's bounding box annotations with tissue-level semantic labels to create duct-level instance masks (details in Section \ref{sec:data}).

\subsection{Semantic segmentation}
\label{sec:sem_seg}
An off-the-shelf segmentation network \cite{mehta2018learning} is applied to generate tissue-level semantic segmentation. The off-the-shelf network splits the input image into non-overlapping patches and predicts a segmentation mask for each patch using a multi-resolution encoder-decoder structure. 

\subsection{Feature Extraction and Classification}
\label{sec:pixel_features}
Several methods (e.g., histogram features, co-occurrence features, and structural features) have been proposed to extract features from tissue-level segmentation masks. 
Deriving from superpixels (regions of similar color with an area), these features allow the encoding of information about tissues (e.g., stromal tissue) and structures (e.g., ducts) present in biopsy images and help improve the diagnostic classification compared to multi-instance learning-based methods \cite{mercan2019patch}.
Histogram frequency features can convey the distribution of tissues in an image, and the co-occurrence features can encode simple spatial relationships. The structure features, extracting frequencies from five layers inside and five layers outside of a duct instance, can capture the changes in the shape of the epithelial structures.

Though the structure features used in \cite{mercan2019assessment} would allow capturing architectural information around the ducts, they are computationally expensive as compared to the histogram and co-occurrence features. 
While computing structure features takes over an hour averagely for each image, computing the proposed features only takes about one second.
The proposed DIOP leverages the duct-, box-, and tissue-level information based on tissue histogram frequency and co-occurrence frequency. This design allows us to replace these computationally expensive features with simple features at three levels. These features are then fed to a classification network (e.g., a random forest or a multi-layer perceptron) to predict the diagnostic category. The ability of our system to aggregate clinically relevant information at different levels allows our framework to outperform existing methods by a significant margin (Section \ref{sec:results}).

\input{duct_rst_fig}

\section{Experimental Results}

\subsection{Dataset}
\label{sec:data}

\noindent \textbf{Diagnostic labels:} The Ductal Instance-Oriented Pipeline was developed by using digital whole slide images created from residual breast biopsy material \cite{oster2013development,weaver2005predicting,carney1996new}.
The dataset consists of a total of 428 ROI images, which were extracted from 240 breast biopsies and categorized by three expert pathologists, who selected the ROIs for each whole slide image and agreed on a consensus diagnosis for each slide. Similar to many previous studies on this dataset \cite{mehta2018net,mercan2019assessment,wu2020mlcd}, we performed diagnosis classification for each of the 428 ROIs into 4 classes: Benign, Atypia, Ductal Carcinoma in Situ, or Invasive Cancer. 
The dataset is unique and is enriched with additional cases in the challenging Atypia and DCIS categories to assist in establishing statistical confidence in accuracy predictions for categorical classification. The enriched image dataset provides additional ROI input data on lower prevalence disease categories for developing the AI pipeline.
The comparison of participants’ diagnoses performance with expert pathologists was previously reported \cite{elmore2015diagnostic}.

\begin{figure}[!h]
    \centering
    \includegraphics[width=0.48\textwidth]{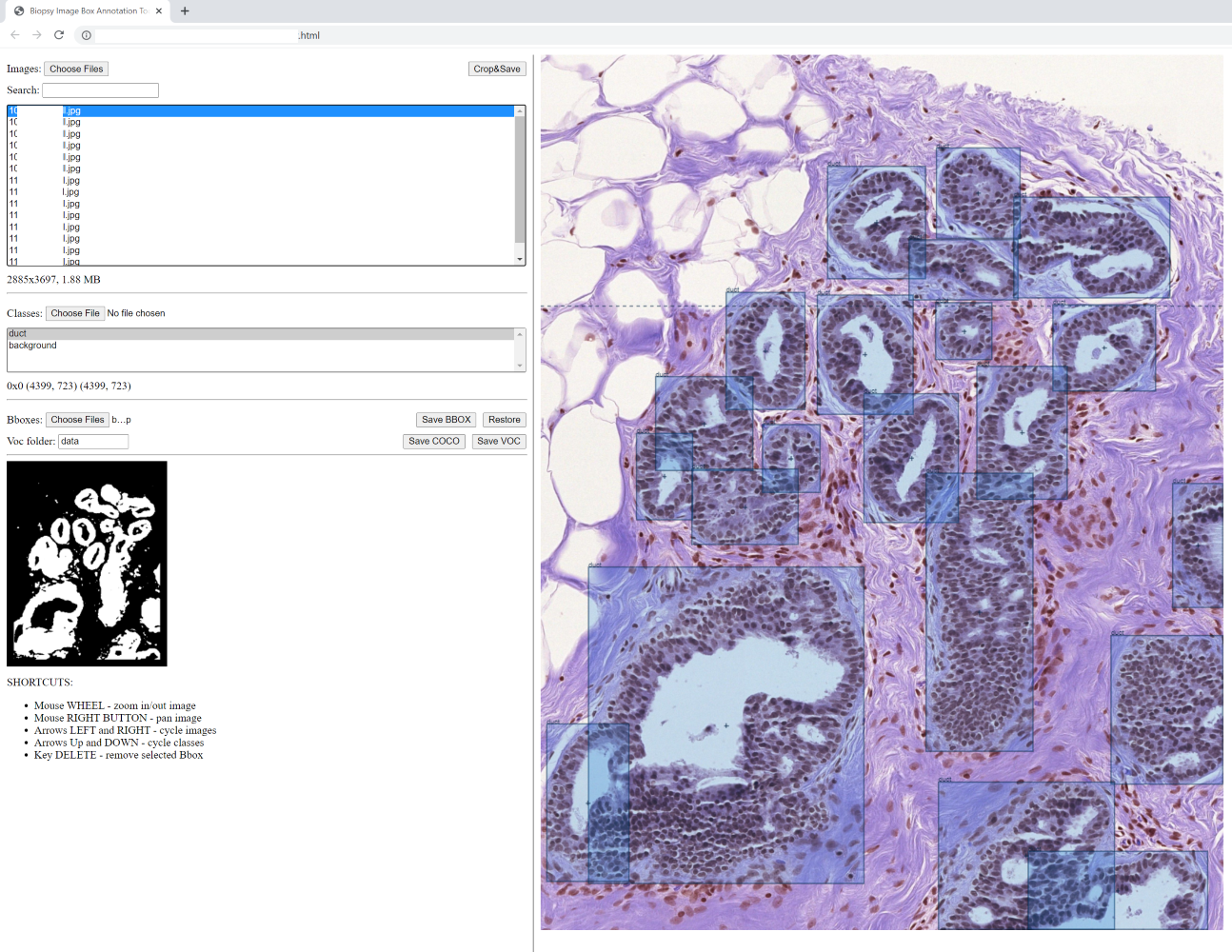}
   \caption[Caption placeholder]{\textbf{The graphical user interface (GUI) to annotate duct instances:} the main panel (right-hand side) shows the input image and allows users to create bounding box annotations. The bottom left side shows the binary tissue mask to guide the annotator. The top left section allows the annotator to load, select, and save images and annotations. Filenames are removed in this visualization for privacy concerns.
   This GUI is developed based on HTML, Javascript, and YOLO BBox Annotation Tool\protect\footnotemark.
   }
   \label{fig:gui}
\end{figure}

\vspace{1mm}
\noindent \textbf{Duct-level instance annotations:} Ductal regions are important bio-markers in diagnosing breast cancer. However, collecting duct-level instance segmentation masks is difficult because pathologists are required to annotate the instances. We created a weakly supervised annotation tool to collect duct-level instance segmentation masks. Our annotation tool is shown in Figure \ref{fig:gui}. We first applied off-the-shelf tissue-level semantic segmentation network to ROIs. Benign epithelium (BE), malignant epithelium (ME), secretion (SC), and necrosis (NC) are tissues that surround ducts. Therefore, we created binary masks by assigning pixels in these tissues as foreground with the remaining pixels as background, as shown in Figure \ref{fig:instance:binary}. These masks were then combined with the bounding box annotations\footnote{The bounding boxes around ducts were marked by an engineering student (Beibin Li) under the supervision of an expert pathologist (Stevan Knezevich).} to produce duct-level instance segmentations. Overall, 4,347 duct instances were marked in 100 ROIs. Note that one bounding box might contain pixels from several ducts, and this annotation strategy can still mistakenly mark pixels from other ducts to the main duct in the bounding box. So, our annotations are only ``silver standard" rather than ``ground truth,"  because they are inexact.

\vspace{1mm}
\noindent \textbf{Tissue-level semantic annotations:} The dataset also provides pixel-wise tissue-level labels for 58 ROIs. An expert pathologist annotated these ROIs into eight tissue classes: background (BG), benign epithelium (BE), malignant epithelium (ME), normal stroma (NS), desmoplastic stroma (DS), secretion (SC), blood (BL), and necrosis (NC). These ROIs were used to train a tissue-level  segmentation model. See \cite{mehta2018learning} for more details.

\input{table_binary}

\subsection{Implementation details}

\noindent \textbf{Duct-level instance segmentation:} We fine-tuned Mask R-CNN\footnote{\url{https://github.com/matterport/Mask\_RCNN}} \cite{he2016deep} (pretrained on the MS-COCO dataset) with ResNet-50 as a backbone network for 30 epochs using SGD with an initial learning rate of 0.01 and a momentum of 0.9. We used duct-level instance segmentation masks produced using our weakly supervised annotation tool for fine-tuning Mask R-CNN. Compared to cellular entities, ductal regions are larger in size and can be easily detected with lower resolution images. Therefore, we resized all ROI mages to a fixed spatial dimension of $512 \times 512$. The dataset was split into an 80:20 ratio: 80 ROIs for training and 20 ROIs for validation. On the validation set, Mask R-CNN achieved a mean intersection over union (mIOU) of 72\% and mean average precision (mAP) of 32\%. Figure \ref{fig:duct_overlap} visualizes duct-level instance segmentation masks produced by Mask R-CNN.

\vspace{1mm}
\noindent \textbf{Tissue-level semantic segmentation:} 
Ductal regions can be identified at lower image resolutions because the shape and texture can help recognition. However, tissue-level segmentation methods do not perform well at lower resolutions, because low-resolution images may lose information about cellular entities, which help differentiate between different tissues. 
Similar to \cite{mercan2019assessment},  the off-the-shelf semantic segmentation method \cite{mehta2018learning} was applied to produce tissue-level semantic masks at x40 magnification.

\vspace{1mm}
\noindent \textbf{Diagnostic classification:} 
we tried a random forest model,  a 3-degree polynomial support vector machine (SVM), an SVM with radial basis function (RBF) kernel, and a multi-layer perceptron (MLP) with four hidden layers (256, 128, 64, and 32 neurons for each layer, similar to \cite{mehta2018net}) for comparison.

\subsection{Classification methods}

Diagnostic classification tasks and baseline methods are introduced below.
We run each experiment 100 times and report the mean performance.

\noindent \textbf{Binary classification:} Emulating the successive decisions made by pathologists, \cite{mercan2019assessment} performed three binary classification tasks (i.e. invasive v.s. non-invasive, atypia \& DCIS v.s. benign, and DCIS v.s. atypia) in their studies.
We performed similar experiments, using leave-one-out cross-validation to evaluate each binary classification model. If the number of features was more than the number of ROIs in a classification task, we performed principal components analysis (PCA) to reduce the number of features to 20 dimensions. We applied a weighted random sampling approach to sample balanced positive and negative samples before training these binary classifiers.

\vspace{1mm}
\noindent \textbf{Multi-class classification:}  We also conducted a 4-way classification experiment to compare our results with previous studies \cite{mehta2018net,mercan2019patch}. We used the same training/validation/testing split as Y-Net \cite{mehta2018net}, an extension of U-net with a separate branch for diagnostic classification; their discriminative masks improved classification accuracy by 7\% over previous feature-engineering methods.
On the other hand, a multi-instance learning based method \cite{mercan2019patch}, analyzing extracted features from a CNN instead of tissue-level semantic information, outperformed previous methods.
We will compare the proposed DIOP with these baseline methods.

\subsection{Main Results}
\label{sec:results}

\vspace{1mm}
\noindent \textbf{Binary classification:} Table \ref{tab:dx_rst} compares the performance of our method with the superpixel and structural features  \cite{mercan2019assessment} in terms of sensitivity, specificity, accuracy, and $F_1$ score. Overall, our method outperforms both methods in all binary classification tasks. We observe that the super-pixel-feature-based method delivers the best performance for the invasive vs. non-invasive task. This is because cancer cells spread out from the ducts in invasive cancer (as shown in Figure \ref{fig:invasive_fail}). This limits both our method and the structure-feature-based method to aggregate information around ducts, resulting in lower performance. In contrast, the super-pixel-feature-based method only accounts for pixel-level information and not structure-level information. Therefore, such methods are resilient to structural changes.

\vspace{1mm}
\noindent \textbf{Multi-class classification:} Table \ref{tab:four_class} compares 4-way classification performance of our method with state-of-the-art methods. Compared to these methods, our method delivers significantly better performance. For example, our method is about 7\% and 3\% more accurate than Y-Net and the multiple instance learning (MIL)-based method. Importantly, our method matches the performance of  pathologists on this dataset.

\begin{table}[t!]
\centering
\begin{tabular}{lr}
\toprule[1.5pt]
\textbf{Method}  &  \textbf{Accuracy}       \\ 
\midrule[1pt]
Pathologists \cite{elmore2015diagnostic} &  0.70  \\
\midrule
MIL with max-pooling \cite{mercan2017multi}  & 0.55   \\ 
MIL with learned fusion \cite{mercan2019patch}   & 0.67     \\ 
Semantic Learning \cite{mehta2018learning}   & 0.55     \\ 
Y-Net \cite{mehta2018net}  &  0.63 \\ 
DIOP (Ours) & \textbf{0.70} $\pm$ 0.02 \\ 
\bottomrule[1.5pt]
\end{tabular}
\caption{Multi-class classification results on the breast biopsy dataset. Our model outperforms existing methods by a significant margin and also, matches the performance of pathologists.}
\label{tab:four_class}
%% we computed accuracy and weighted $F_1$ for our model, where the accuracy of our system is shown in the table with STD=1.84\%, and the weighted $F_1$ score reaches the mean of 68.7\% (STD=1.95\%).
\end{table}

\input{ablation_tables}

\subsection{Ablations}
To understand the components of our system in detail, we perform the following experiments:

\vspace{1mm}
\noindent \textbf{Impact of duct-level instance segmentation and tissue-level semantic segmentation:} Following \cite{mercan2017digital}, we extracted L*a*b, hematoxylin and eosin (H\&E), and local binary pattern (LBP) histogram features for the duct-only method. For the other two methods (tissue-only and tissue+duct), we extracted histogram and co-occurrence tissue-level features (similar to \cite{mehta2018net,mehta2018learning}). Table \ref{tab:ablation_seg} shows that the method that uses both duct- and tissue-level information delivers the best performance.
 
\begin{table}[b!]
    \centering
    \begin{tabular}{ll|c}
    \toprule[1.5pt]
     \textbf{Ducts} & \textbf{Tissue} & \textbf{Accuracy} \\
     \midrule[1pt]
     \checkmark &     & 0.57 \\
      &  \checkmark  & 0.67 \\
     \checkmark &  \checkmark  & \textbf{0.70} \\
     \bottomrule[1.5pt]
    \end{tabular}
    \caption{Impact of duct-level instance segmentation and tissue-level semantic segmentation.}
    \label{tab:ablation_seg}
\end{table}

\vspace{1mm}
\noindent \textbf{Impact of different features:} To aggregate the information about different structures present in the breast biopsy images, previous methods have proposed different features, namely structure features, histogram features, and co-occurrence features. Table \ref{tab:ablation_features} compares the performance of our method with these different features. Using both histogram and co-occurrence features delivers the best performance. We note that our method delivers a good performance with co-occurrence features alone. This is because the co-occurrence matrix encodes strong relationships between different tissues.

\begin{table}[t!]
    \centering
    \begin{tabular}{llr}
\toprule[1.5pt]
\textbf{Histogram} & \textbf{Co-occurrence} & \textbf{Accuracy} \\
\midrule[1pt]
\checkmark & & 0.66 \\
 & \checkmark & 0.69 \\
\checkmark & \checkmark & \textbf{0.70} \\
\bottomrule[1.5pt]
\end{tabular}
    \caption{This table studies the impact of different features extracted from duct-level and tissue-level masks (Section \ref{sec:pixel_features}). We did not use superpixel and structural features \cite{mercan2017digital} because (1) they are computationally expensive and (2) our method delivers better performance with these simple features (\textbf{0.70} vs. 0.66).}
    \label{tab:ablation_features}
\end{table}

\vspace{1mm}
\noindent \textbf{Impact of extracting features from different levels:} Our framework in Section \ref{sec:system} encodes information from three different levels: (1) tissue-level segmentation mask for the whole image, (2) tissue-level mask for duct bounding boxes and (3) tissue-level mask for duct instances. Table \ref{tab:ablation_levels} shows that extracted features from all levels help improve performance. 

\begin{table}[t!]
    \centering
    \begin{tabular}{lr}
    \toprule[1.5pt]
    \textbf{Method} & \textbf{Accuracy} \\ 
        \midrule[1pt]
        Tissue in ROI & 0.67 \\
        Tissue in Duct box  & 0.66 \\
        Tissue in Duct mask  & 0.69  \\
        Tissue in Duct mask + ROI  & 0.69  \\
        Tissue in Duct box + ROI  & 0.67   \\
        Tissue in Duct box + mask &  0.69  \\
        Tissue (All) &  \textbf{0.70}   \\
    \bottomrule[1.5pt]
    \end{tabular}
    \caption{Impact of extracting features from different levels (segmentation ROI mask, duct mask, and duct boxes).}
    \label{tab:ablation_levels}
\end{table}

\vspace{1mm}
\noindent \textbf{Impact of classifiers:} We study the impact of different classifiers in Table \ref{tab:ablation_classifier}. Compared to widely used MLP and SVM, the random forest delivered the best performance. This is likely because random forests reduce high-variance by ensembling many trees into one model. This reduces over-fitting and improves performance, especially on small datasets (like ours).

\begin{table}[t!]
    \centering
    \begin{tabular}{lr}
        \toprule[1.5pt]
        \textbf{Method} & \textbf{Accuracy} \\ 
        \midrule[1pt]
        SVM (polynomial) & 0.62 \\
        SVM (RBF-kernel) & 0.65 \\
        MLP  & 0.66  \\ 
        Random Forest (DIOP) &  \textbf{0.70}  \\ 
        \bottomrule
    \end{tabular}
    \caption{Impact of different classification algorithms.}
    \label{tab:ablation_classifier}
\end{table}

\section{Discussion}
\label{sec:discussion}

% \subsection{Interpretation}
\textbf{Interpretation:}
While parameters inside a random forest can be hard to understand, we adapted SHAP \cite{lundberg2017unified}, a game theory-based approach, for interpretation.
After training our diagnosis model, we applied SHAP to interpret the diagnostic decision for each ROI and search for the most important features among all ROIs.
Table \ref{tab:shap} compares the 10 most important features from the DIOP and from the tissue-level machine learning model (with 0.67 accuracy) in the ablation experiments.

The BD (boundary of ducts) values in co-occurrence features occur when a pixel is adjacent to the border of a mask or bounding box, which matches the boundary of duct tissues and thus does not have a second pixel to co-occur with.
DIOP identifies two co-occurrence features related to BD as important features, which is consistent with the intuition of structure features.
Even if the tissue-level model can identify similar co-occurrence features, it is unable to use information inside ductal regions;
on the other hand, DIOP mostly focuses on duct masks and also bounding boxes, and it only ranks one ROI-level feature to the top-10 most important features. 
More clinical studies are needed to verify the consistency of our interpretations with pathologists in the future.

\begin{table}[t!]
\begin{tabular}{l|c|c}
\toprule[1.5pt]
Rank & \textbf{DIOP (ours)} & \textbf{Tissue-level model} \\
\midrule[1pt]
1 & BD \& BE in duct mask                  & ME \& NC in ROI \\
2 & ME \& NC in duct mask                   & BG \& NC in ROI \\
3 & BD \& NC in duct mask                  & SC freq in ROI  \\
4 & BE \& NS in bounding box                & BE freq in ROI  \\
5 & BG \& NC in duct mask                   & BE \& SC in ROI \\
6 & BE \& SC in ROI                         & ME \& NS in ROI \\
7 & ME \& SC in bounding box                & BE \& NS in ROI \\
8 & NC freq in bounding box  & NC freq in ROI  \\
9 & BE \& SC in bounding box                & NS \& NC in ROI \\
10 & DS freq in duct mask & SC \& NC in ROI\\
\bottomrule[1.5pt]
\end{tabular}
\caption{The top-10 important features from SHAP interpretation. \textbf{Left:} results from the Ductal Instance-oriented Pipeline (DIOP). \textbf{Right:} results from the classifier for tissue-level semantic features (aka ROI-level features).}
\label{tab:shap}
\end{table}

\textbf{Diagnosis:} 
The proposed DIOP outperforms existing end-to-end and feature engineering approaches.
For the four-way classification task, the DIOP achieves comparable performance to general pathologists.
In Section \ref{sec:results}, the general pathologists' diagnostic accuracy is 70\% for this unique dataset, which over-sampled DCIS and Atypia cases. 
In the real-world setting, pathologists diagnosing accuracy is over 92\% \cite{elmore2016variability}.

\textbf{Limitation:}
Note that our dataset was only obtained from 240 biopsies, and additional studies are needed to fully examine our method.
% We used the same segmentation models as in Mercan et al. \cite{mercan2019assessment} for a fair comparison. 
% However, Y-Net \cite{mehta2018net} and MLCD \cite{wu2020mlcd} were trained on a different training/validation/testing split for the segmentation task, making comparison imperfect. 
With future improvement in semantic segmentation and instance segmentation approaches, our system has the potential to achieve higher accuracy.  
In this study, we explored clinical relevant features for cancer diagnosis, but we used two separate networks for duct-level and tissue-level segmentation; these could be combined in the future.

\textbf{Data Annotation:}
In this study, AI-generated semantic segmentation masks helped the annotator to understand breast biopsies and to perform some easy annotation tasks.
These annotations were then used to train instance segmentation models for diagnosis purposes.
More comprehensive studies, such as controlled and counterbalanced human factor experiments, are needed to investigate the effectiveness of this human-in-the-loop design and other types of human-AI interaction, such as educating pathologist trainees.

\textbf{Future directions:}
Many questions need to be answered before deploying these CAD systems, for examples:
the generalizability to different datasets, 
the interpretability of models, 
the robustness of application to diverse images sets,
and the vulnerability to noise and adversarial attacks.
Abnormal breast histopathology and breast cancer are complex and heterogeneous disease processes that still require human experts to supervise the important diagnostic decision process.

\section{Conclusion}
In this study, we proposed the Ductal Instance-Oriented Pipeline for breast pathology and cancer diagnosis, which contains a duct-level instance segmenter, a tissue-level semantic segmenter, three-levels of pixel-wise features, and a diagnostic classifier.
To train the special instance-level segmentation model, we adapted weak annotation and human-in-the-loop design to acquire training data.
The proposed method outperforms previous computer-aided approaches in all diagnostic tasks.
It also outperforms general pathologists in 2 out of 3 binary classification tasks and almost matches overall performance for general pathologists on our unique dataset.
More comprehensive studies are needed to validate our approach in the future.

\footnotetext{https://github.com/drainingsun/ybat}

\small
\bibliographystyle{unsrt} % sort the reference order by appearance
\bibliography{main}
\raggedbottom % Do not stretch the reference to fill a full page.

\end{document}

%% file: duct_rst_fig.tex
\begin{figure*}[t!]
    \centering
    \begin{subfigure}[b]{0.48\columnwidth}
        \centering
        \includegraphics[width=\columnwidth, height=100px]{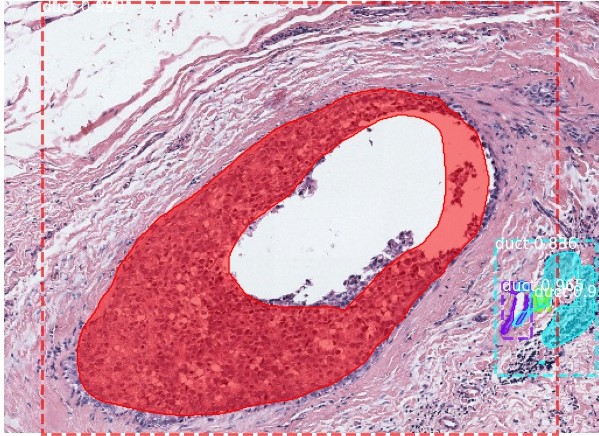}
        \caption{}
    \end{subfigure}
    \hfill
    \begin{subfigure}[b]{0.48\columnwidth}
        \centering
        \includegraphics[width=\columnwidth, height=100px]{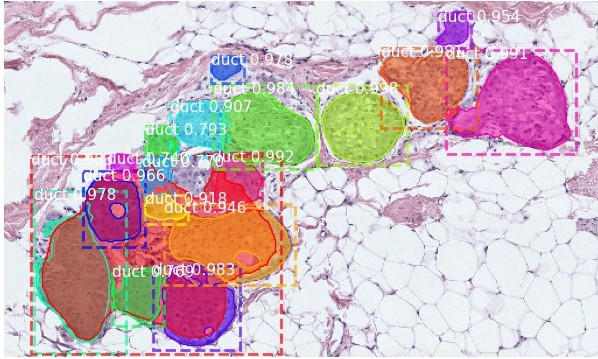}
        \caption{}
    \end{subfigure}
    \hfill
    \begin{subfigure}[b]{0.48\columnwidth}
        \centering
        \includegraphics[width=\columnwidth, height=100px]{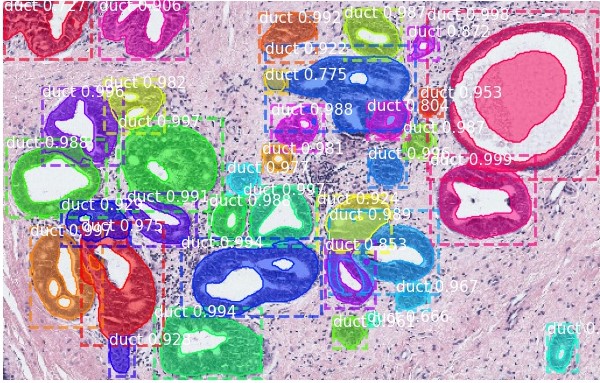}
        \caption{}
    \end{subfigure}
    \hfill
    \begin{subfigure}[b]{0.48\columnwidth}
        \centering
        \includegraphics[width=\columnwidth, height=100px]{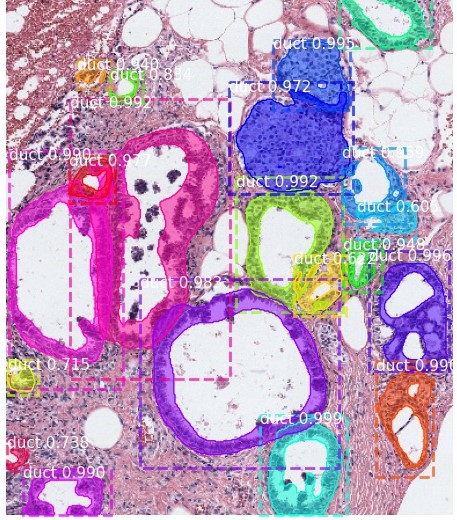}
        \caption{}
    \end{subfigure}
    \vfill
    \begin{subfigure}[b]{0.48\columnwidth}
        \centering
        \includegraphics[width=\columnwidth, height=100px]{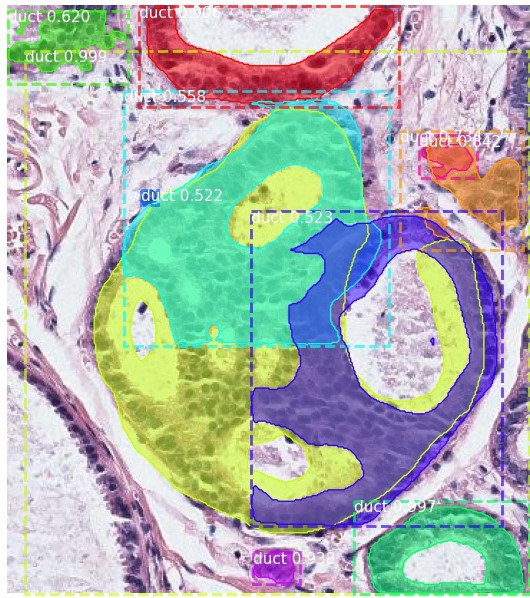}
        \caption{}
        \label{fig:duct_good_overlap}
    \end{subfigure}
     \hfill
    \begin{subfigure}[b]{0.48\columnwidth}
        \centering
        \includegraphics[width=\columnwidth, height=100px]{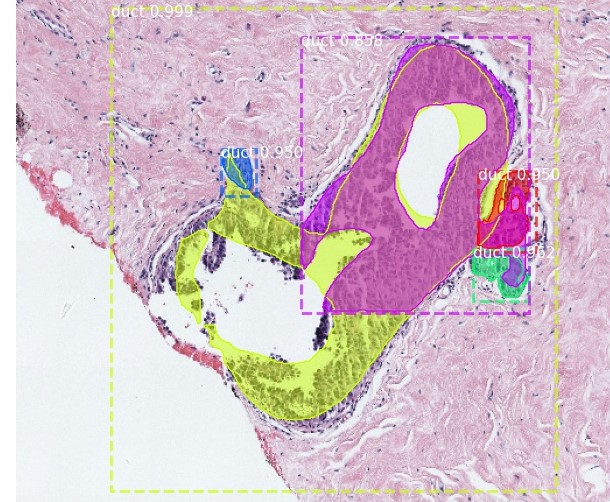}
        \caption{}
        \label{fig:duct_overlap}
    \end{subfigure}
     \hfill
    \begin{subfigure}[b]{0.48\columnwidth}
        \centering
        \includegraphics[width=\columnwidth, height=100px]{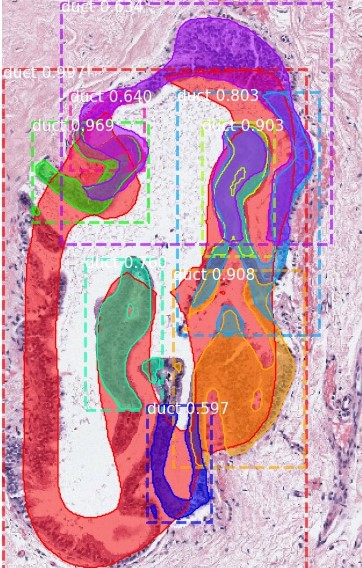}
        \caption{}
        \label{fig:duct_cutted}
    \end{subfigure}
     \hfill
    \begin{subfigure}[b]{0.48\columnwidth}
        \centering
        \includegraphics[width=\columnwidth, height=100px]{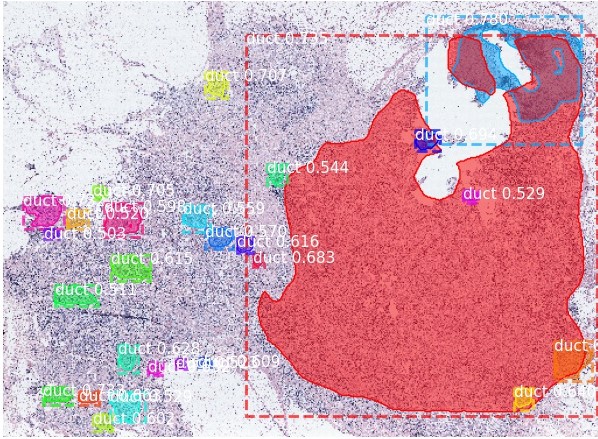}
        \caption{}
        \label{fig:invasive_fail}
    \end{subfigure}
  \caption{\textbf{Ductal segmentation testing result:} each color represents one instance of a duct in the biopsy. 
  The top row shows four examples with satisfiable duct identification results, and the bottom row shows four imperfect examples.
  Examples (e), (f), and (g) have taken a single irregular and expanded duct and split it into multiple duct structures.
  Cancerous cells have escaped from ducts in (h), and our system mistakenly marks a big region (in red) as one duct.
  % In \ref{fig:duct_overlap}, the duct marked in yellow overlapped too much with the duct marked in pink. 
  }
  \label{fig:duct_rst}
\end{figure*}

%% file: table_binary.tex
\begin{table*}[t!]
\centering
\begin{tabular}{l|l|rrrr}
\toprule[1.5pt]
\textbf{Task}                      & \textbf{Features} & \textbf{Sensitivity}    & \textbf{Specificity} & \textbf{Accuracy}    & \textbf{F$_1$}          \\
\midrule[1pt]
\multirow{4}{*}{\textbf{Invasive vs Non-invasive}}   & \textit{Pathologists} \cite{elmore2015diagnostic} & \textit{0.84}  & \textit{0.99}  & \textit{0.98}  & \textit{0.86}  \\
  & Superpixel Features \cite{mercan2019assessment}                & \textbf{0.70}                            & 0.95                                 & 0.94                                 & 0.62                              \\
  & Structure Features  \cite{mercan2019assessment}                 & 0.49                                      & 0.96              & 0.91               & 0.51                \\
 & Duct-RCNN (Ours)              & 0.62                 & \textbf{0.98}        & \textbf{0.95}       & \textbf{0.73}       \\
\midrule
\multirow{4}{*}{\textbf{Atypia and DCIS vs Benign}} & \textit{Pathologists} & \textit{0.72} & \textit{0.62} & \textit{0.81} & \textit{0.51} \\
 & Superpixel Features                & 0.79                         & 0.41                                & 0.70                             & 0.46                                \\
& Structure Features & \textbf{0.85}                & 0.45             & 0.70              & 0.50               \\
& Duct-RCNN  (Ours)  & \textbf{0.85} & \textbf{0.63}  & \textbf{0.79}  & \textbf{0.59}       \\
\midrule
 \multirow{4}{*}{\textbf{DCIS vs Atypia}}                                                    & \textit{Pathologists}              & \textit{0.70}                       & \textit{0.82}                      & \textit{0.80}                     & \textit{0.76}                       
                                                    \\
           & Superpixel Features                & 0.88                                     & 0.78                                 & 0.83                                 & 0.86                                 \\
                                                    & Structure Features                 & 0.89                   & 0.80               & 0.85               & 0.87               \\
                                                    & Duct-RCNN (Ours)             & \textbf{0.91} & \textbf{0.89} &       \textbf{0.90}      & \textbf{0.92}      \\
 \bottomrule[1.5pt]
\end{tabular}
\caption{\textbf{Diagnosis results for binary classification:} we show sensitivity, specificity, accuracy, and $F_1$ score for all models.
We highlight the best machine performances in this table, and pathologists' performances are provided for comparison. 
The results for superpixel features and structure features are from \cite{mercan2017digital,mercan2019assessment}, where the standard deviations (STD) are not reported.
}
\label{tab:dx_rst}
\end{table*}

%% file: ablation_tables.tex
\begin{comment}
\begin{table}[b]
    \centering
    \begin{tabular}{lr}
\toprule
                                   \textbf{Method} & \textbf{Accuracy} \\ 
\midrule
 Tissue-level Segmentation                                & 0.67                                     \\
 Duct-level Segmentation                                       &  0.57                                          \\ 
 Tissue- and Duct-level Segmentation           &   \textbf{0.70}                                       \\ 
                                                      \bottomrule
\end{tabular}
    \caption{Ablation experiment results}
    \label{tab:ablation:seg}
\end{table}
\end{comment}